\documentclass[letterpaper, 10pt, conference]{ieeeconf}  

\IEEEoverridecommandlockouts                              

\usepackage{amsmath} 
\usepackage{amssymb}  
\usepackage{mathrsfs}  

\usepackage{adjustbox}
\usepackage{subcaption}
\usepackage{url}
\usepackage{xcolor}
\usepackage[pdfencoding=auto]{hyperref}
\usepackage[normalem]{ulem} 

\hypersetup{
    breaklinks   = true, 
    colorlinks   = true, 
    linkcolor = black,
    urlcolor  = black,
    citecolor = black,
    anchorcolor = black,
    pdfstartview = {FitV}, 
    bookmarksnumbered  = true, 
    bookmarksopen      = true, 
    bookmarksopenlevel = 0, 
    bookmarkstype      = toc, 
    pdfpagemode        = {UseOutlines}, 
}

\usepackage[ruled, noend, linesnumbered]{algorithm2e}
\SetAlgorithmName{Pseudo.}{}{}

\title{\LARGE \bf
Online path planning for kinematic-constrained UAVs in a dynamic environment based on a Differential Evolution algorithm
}

\author{Elias J. R. Freitas$^{1}$,  Miri Weiss Cohen$^{2}$, Frederico G. Guimar\~aes$^{3}$ and  Luciano C. A. Pimenta$^{4}$
\thanks{
This work was carried out with the support of the Instituto Federal de Educação, Ciência e Tecnologia de Minas Gerais – Campus Ibirité. This study was financed in part by the Coordenação de Aperfeiçoamento de Pessoal de Nível Superior - Brasil (CAPES) - Finance Code 001. This work was also supported in part by the project INCT (National Institute of Science and Technology) under the grant CNPq (Brazilian National Research Council) 465755/2014-3 and FAPESP (São Paulo Research Foundation) 2014/50851-0, CNPq (grant numbers 309925/2023-1 and 407063/2021-8), and Fundação de Amparo à Pesquisa do Estado de Minas Gerais (FAPEMIG), under grant number APQ-0063023. 
}
\thanks{$^{1}$Elias José de Rezende Freitas is with the Universidade Federal de Minas Gerais, UFMG, in the Graduate Program in Electrical Engineering. Av. Antônio Carlos 6627, 31270-901. Belo Horizonte, MG, Brazil. {\tt\small elias.freitas@ifmg.edu.br}}
\thanks{$^{2}$Miri W. Cohen is with Braude College of Engineering, 
Karmiel, Israel.}%
\thanks{$^{3,4}$ Frederico G. Guimar\~aes and Luciano C. A. Pimenta are with Universidade Federal de Minas Gerais, UFMG, Brazil.}%
}

\begin{document}

\maketitle
\thispagestyle{empty}
\pagestyle{empty}

\begin{abstract}

This research presents an online path planner for Unmanned Aerial Vehicles (UAVs) that can handle dynamic obstacles and UAV motion constraints, including maximum curvature and desired orientations. Our proposed planner uses a NURBS path representation and a Differential Evolution algorithm, incorporating concepts from the Velocity Obstacle approach in a constraint function. Initial results show that our approach is feasible and provides a foundation for future extensions to three-dimensional (3D) environments.


\end{abstract}


\section{INTRODUCTION}

The increasing use of fixed-wing Unmanned Aerial Vehicles (UAVs) is driven by several factors, such as long-range, high speeds, and superior payload capacity compared to quadrotors.  Combined with motion planning strategies, these advantages enable fixed-wing UAVs also to navigate in complex environments, avoiding obstacles safely.


Motion planning involves both path planning and collision avoidance strategies. Typically, a global planner uses path planning techniques to navigate a known environment and combines these with a local planner that reacts to avoid collisions with dynamic or new obstacles. Additionally, fixed-wing UAVs present unique challenges due to motion constraints, such as maximum curvature, maximum/minimum climb angle, and desired orientations (yaw and pitch angles), making it difficult to find a feasible path and avoid collisions. To address this, some research papers have proposed Dubins-based path planners, which ensure optimal path length in obstacle-free environments~\cite{park2022three}\cite{vavna2020minimal}, while others have suggested combined strategies like using a planning-based algorithm with Model Predictive Control (MPC)~\cite{primatesta2021mp_sampling} or with Velocity Obstacle approach (VO)~\cite{zhang2020collision} to facilitate fixed-wing navigation.

In our recent work~\cite{freitas2024_de3dnurbs}, we proposed a novel Differential Evolution-based path planner that handles kinematic-constrained UAVs. In this approach, we also show that using the Non-Uniform Rational B-spline (NURBS) curve as the path representation can provide a more flexible planner than using the B-spline representation. In addition, our approach does not require additional smoothing steps as other approaches do for fixed-wing UAVs~\cite{chai2022multi}. Planning the motion to avoid dynamic obstacles remains a challenging open problem. Simply adding a local planner can guide the fixed wing to a local minimum or unfeasible situation.

\begin{figure}
    \centering
    \includegraphics[width=1\linewidth]{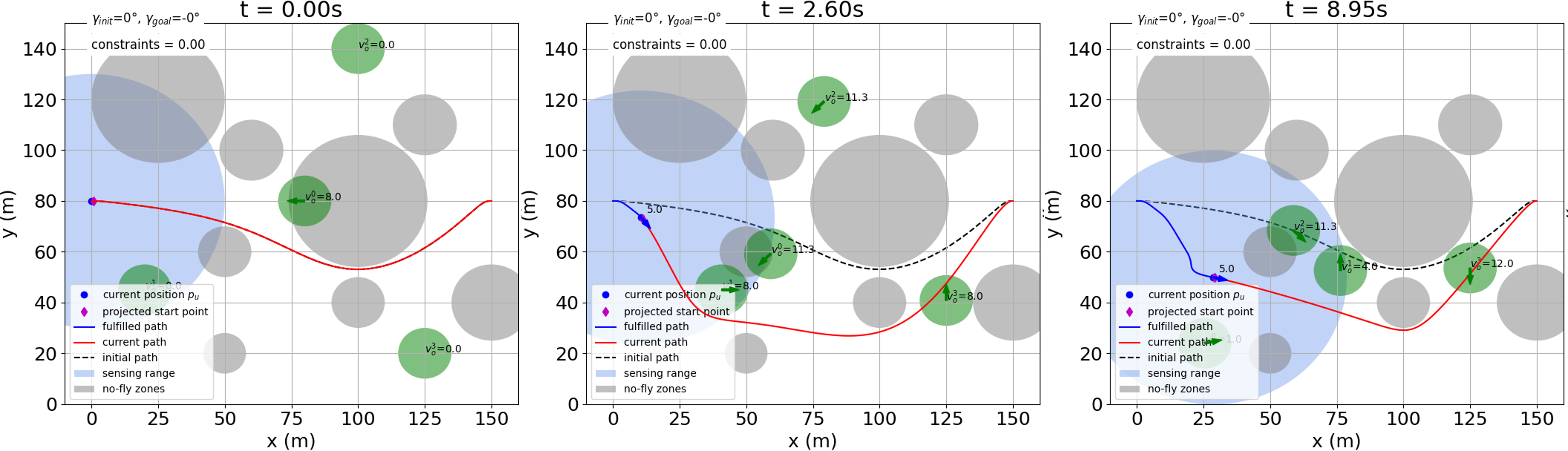}
    \caption{Snapshots of the deliberative planning process. The path fulfilled up to that point is in blue and the planned curve is in indigo. 
    }
    \label{fig:overview}
\end{figure}

The primary focus of this work is to expand our earlier formulation from an offline to an online planner. We have developed a new online path planner that takes into account dynamic obstacles, in addition to the motion constraints of the UAVs, and seamlessly incorporates them into the deliberative planning process.

\section{PROBLEM STATEMENT}

Consider that a high-level mission planner determines a sequence of $N_w$ desired positions and orientations, $W=\{W_1 ,..., W_{N_w}\}$, which can serve, for example, as reference waypoints for inspection, data collection, and surveillance missions for a fixed-wing UAV. Our problem is to provide an online and deliberative \emph{feasible path} that satisfies motion constraints for fixed-wing UAVs in a dynamic environment at each re-planning interval ($t_s$). 

A feasible path refers to a curve \(\mathit{C}(s)\), where \(s\) is a real parameter. This curve must reach the specified waypoints in the given order while adhering to the constraints. Additionally, it should steer clear of previously known obstacles and any unknowns detected within a perception range that may be moving in the environment.


In this study, we examine a scenario in which the UAV is flying steadily at a constant speed, meaning that its thrust force equals the drag force. This means that the UAV cannot change its speed during this flight regime, but it can change direction. This adds complexity to the motion that previous works~\cite{zhang2020collision}\cite{xu2023shunted}\cite{liu2024velocity} have not addressed.

For early results, we assume constant altitude; however, the proposed method can be extended to 3D scenarios. The UAV kinematic model is represented by a Dubins model, and the path planning problem can be solved in 2D space, considering the maximum curvature ($\kappa_{\mathrm{max}}$) and the desired yaw orientations ($\gamma_{\mathrm{init}}$, $\gamma_{\mathrm{goal}}$). 


    

We can navigate (i) unknown environments and (ii) initially known environments $(\Omega)$, which contain restricted spaces with no-fly (or threatening) zones $(\Omega_{\mathrm{obs}})$. These zones may include areas of electromagnetic interference, radar detection zones, or areas vulnerable to artillery attack~\cite{chai2022multi}. We assume that changes and/or the presence of new obstacles in the environment can be detected within a limited sensing range ($R_{\mathrm{view}}$), returning, with the necessary precision, the position and velocity vector of such objects at each interval $t_s$.
Considering a safe radius ($R_{\mathrm{safe}}$), the center of the robot on the curve must be kept in the free space $\Omega_{\mathrm{free}}$ throughout the entire mission.

\section{METHODS}

The deliberative planning process is summarized in the Pseudocode~\ref{alg:anytime}. The input of our proposed method is an initial path, for example, if we have no information about the environment, the path can be a straight line to each pair of configurations, or if an initial environment is known, an initial path that satisfies the kinematic constraints can be obtained by our DE-NURBS approach~\cite{freitas2024_de3dnurbs}. The path is represented by a Non-Uniform Rational B-Spline (NURBS) curve~\cite{pieg1996lnurbs}, consisting of $N_p$ control points ($\mathbf{P}$) and for each control point an associated weight ($\mathbf{w}$). To get the desired orientation, we add three collinear points to the first and last control points, sized by the factors $\lambda_1$ and $\lambda_2$.



\begin{algorithm}[t]
\caption{Deliberative planning process}
\label{alg:anytime}
\SetAlgoLined
\footnotesize 

\textbf{Input}: Initial control points ($\mathbf{P}$) and weights ($\mathbf{w}$) of the NURBS path $C(s)$ from $W_i$ to $W_{i+1}$, and the re-planning interval $t_s$
   
\textbf{Output}: Feasible path from $\Delta\chi^* = [ \Delta\mathbf{P}, \Delta\mathbf{w}, \lambda_1, \lambda_2 ]$

\While{not termination criteria}{

Update the local map (inside a sensing range)

Compute the projection point $\mathbf{p_u}(t_s)$ in $C(s)$ using \cite{rezende2022constructive}

Cut the path to make $C(0) = \mathbf{p_u}(t_s)$

Solve \eqref{eq:opt_problem}, considering $VO_{o_i/u}^{\tau}$

\While{not solved \eqref{eq:opt_problem} }{ 

Track the previous curve using a vector field \cite{rezende2022constructive}

}

Update the path based on the optimal variation change $\Delta\chi^*$ in the parameters of the $C(s)$

}

\end{algorithm}



Unlike our previous approach~\cite{freitas2024_de3dnurbs}, considering the flexibility of the NURBS curve, we do not optimize the parameters of the curve directly, but a variation of these parameters  ($\Delta\chi$) from an initial path, which allows a faster convergence. In this study, we use the energy cost of the flight, 
which is directly proportional to the length of the path, keeping the speed constant and ignoring the wind. Then, our optimization problem can be defined as:
\begin{align}
\begin{split}
    & \Delta\chi^* = \arg\min f = \int_{0}^1 C(s)ds+ || \Delta\mathcal{P}||, \\
     \text{s.t.:} \\
    & \mathbf{p_u} \notin \Omega_{\mathrm{obs}} , \\
     & \kappa(s) \leq \kappa_{\mathrm{max}}, \\
    & \mathbf{v_u}(s) \notin VO_{o_i/u}^{\tau}~\forall i \in [1, N_\mathrm{obs}],
\end{split}
\label{eq:opt_problem}
\end{align}
where $\mathbf{p_u}(s) $ is the position of the UAV along the path, in which $\mathbf{p_u}(0)$ is the projected current position in the re-planning interval $t_s$ based on the previous path and using the curve tracking control~\cite{rezende2022constructive}; $\kappa_{\mathrm{max}}$ is inversely proportional to the minimum turning radius ($\rho_{\mathrm{min}}$) that the fixed wing can achieve in space; $\kappa(s)$ is the curvature evaluated on the path from $s \in [0,1]$.
From the Velocity Obstacle approach~\cite{van2008reciprocal}, it's possible to check potential collisions by determining if the relative velocity of all $N_{\mathrm{obs}}$ obstacles relative to the UAV along the curve falls within the velocity space $VO_{{o_i}/u}^{\tau} = \{ \mathbf{v_u} \mid \mathbf{v_u} - \mathbf{v}_{o_i} \in RCC_{r_{o_i} + r_u}^{\tau} \}$, 
%
where \( RCC_{r_{o_i} + r_u}^{\tau} \) represents the cone of potential collision during an interval of $\tau$ seconds, given the position, velocity and the radius of the fixed-wing UAV ($\mathbf{p_u}, \mathbf{v_u}, r_u$) and the $i$th obstacle ($\mathbf{p_{o_i}}, \mathbf{v_{o_i}}, r_{o_i} = R_{\mathrm{safe}}$). This constraint is observed for $s \in [0, s_{\tau}]$, where $s_{\tau}$ is the projected position of the robot at time $\tau$.

To solve this problem at each interval $t_s$, we apply the Linear-Success-History Based Adaptive Differential Evolution for Constrained Optimization Problems (LSHADE-COP) algorithm, which can produce solutions that converge to a feasible optimal solution better than other nature-inspired algorithms~\cite{freitas2024_de3dnurbs}. In our novel proposal, the Velocity Obstacle is considered for each individual in the algorithm population, which gets a possible variation in the initial curve and penalizes future collisions with the detected obstacles. In this way, the best feasible safe path is obtained.


Since our planner provides a curve at least $k$ (degree of the NURBS curve) differentiable on $s$, we can use the approach in \cite{rezende2022constructive} to map the curve path into a vector field control. For that, we compute a vector field $\Psi(\mathbf{p_u}):\mathbb{R}^3\rightarrow\mathbb{R}^3$, where $\mathbf{p_u} \in \mathbb{R}^3$ is a point so that converge to and follow the curve $\mathit{C}(s)$.

\section{SIMULATION RESULTS}





We performed different simulations, changing the number of obstacles in the environment and their velocities. Fig.~\ref{fig:overview} shows snapshots of the planning process in a dynamic environment. We also compared our approach with the re-planning approach without the Velocity Obstacle constraint. The results show that our proposal provided safe navigation even in a crowded environment. Besides that, we can consider a complete mission, where our planner changes only a section of the initial planned NURBS curve to avoid dynamic obstacles. More details are in the supplementary video: \url{https://youtu.be/bZY4cy_YA6s}.


\section{CONCLUSIONS}

In this paper, we proposed an online planner that provides feasible paths, considering kinematic constraints, and it can operate with a re-planning interval of $t_s\approx0.1$ seconds, making it about 12,000 times faster than our previous planner~\cite{freitas2024_de3dnurbs}. Furthermore, we introduced a new constraint into the optimization problem based on the Velocity Obstacle approach. This allows for the avoidance of dynamic obstacles within a limited sensing range.

\bibliographystyle{IEEEtran}
\bibliography{references}

\end{document}